\documentclass[11pt,a4paper]{article}
\usepackage[hyperref]{emnlp2018}
\usepackage{amsmath}
\usepackage{enumitem}
\usepackage{graphics}
\usepackage{latexsym}
\usepackage{multirow}
\usepackage{times}
\usepackage{url}
\usepackage{xspace}

\aclfinalcopy 

\newcommand\ROUGE{\textsc{Rouge-L}\xspace}
\newcommand\NIST{\textsc{Nist}\xspace}
\newcommand\BLEU{\textsc{Bleu}\xspace}

\setlist[itemize]{leftmargin=*}

\title{Unsupervised Natural Language Generation with Denoising Autoencoders}

\author{Markus Freitag \and Scott Roy \\
  Google AI\\
  {\tt \{freitag,hsr\}@google.com} \\}

\date{}

\begin{document}
\maketitle
\begin{abstract}
Generating text from structured data is important for various tasks such as question answering and dialog systems.
We show that in at least one domain, without any supervision and only based on unlabeled text, we are able to build a Natural Language Generation (NLG) system
with higher performance than supervised approaches. 
In our approach, we interpret the structured data as a corrupt representation of the desired output and use a denoising auto-encoder to reconstruct the sentence. 
We show how to introduce noise into training examples that do not contain structured data, and that the resulting denoising auto-encoder generalizes to generate correct sentences when given structured data.
\end{abstract}

\section{Introduction}
\label{sec:introduction}
Natural Language Generation (NLG) is the task of generating text from structured data.
Recent success in Deep Learning motivated researchers to use neural networks instead of
human designed rules and templates to generate meaningful sentences from structured information.
However, these supervised models work well only when provided either massive amounts of labeled data or when restricted to a limited domain.
Unfortunately, labeled examples are costly to obtain and are non-existent for many domains. Conversely, large amount of unlabeled data are often freely available
in many languages.

A labeled example is given in Table~\ref{tab:training_example}.
One labeled example consists of a set of slot pairs and at least one golden target sequence. Each slot pair has a slot name (e.g. "name") and a slot value (e.g. "Loch Fyne").
In this work, we present an unsupervised NLG approach that learns its parameters without the slot pairs on target sequences only.
We use the approach of a denoising auto-encoder (DAE) \cite{vincent2008extracting} to train our model.
During training, we use corrupt versions of each target sequence as input and learn to reconstruct the correct sequence
using a sequence-to-sequence network \cite{kalchbrenner+blunsom:2013,sutskever+:2014,bahdanau+:2014}. We show how to introduce noise into the training data in such a way that
the resulting DAE is capable of generating sentences out of a set of slot pairs. Taking advantage of using unlabeled data only, we also incorporate out-of-domain data into the training process to improve the quality of the generated text.

\begin{table}[!t]
    \centering
    \setlength\tabcolsep{4pt}
    \begin{tabular}{c|c|c|c}
        name & type & food & family friendly \\ \hline
        Loch Fyne & restaurant  & Indian & yes
    \end{tabular}
    \caption{Three possible correct target sequences for the structured data above: 
(a) There is an Indian restaurant that is kids friendly. It is Loch Fyne.
(b) Loch Fyne is a well-received restaurant with a wide range of delicious Indian food. It also delivers a fantastic service to young children.
(c) Loch Fyne is a family friendly restaurant providing Indian food.}
    \label{tab:training_example}
\end{table}

\section{Network}
\label{sec:network}
In all our experiments, we use our in-house attention-based sequence-to-sequence (seq2seq) implementation which is similar to ~\newcite{bahdanau+:2014}. 
The approach is based on an encoder-decoder network.
The encoder employs a bi-directional RNN to 
encode the input words ${\bf{x}}=({x_1, ... , x_l})$ 
into a sequence of hidden states ${\bf{h}}=({h_1, ..., h_l})$, where $l$ is the length of the input sequence.
Each $h_i$ is a concatenation of a left-to-right $\overrightarrow{h_i}$
and a right-to-left $\overleftarrow{h_i}$ RNN:
\[
h_{i} = 
\begin{bmatrix}
\overleftarrow{h}_i \\ 
\overrightarrow{h}_i \\
\end{bmatrix}
=
\begin{bmatrix}
\overleftarrow{f}(x_i, \overleftarrow{h}_{i+1}) \\
\overrightarrow{f}(x_i, \overrightarrow{h}_{i-1}) \\
\end{bmatrix}
\]
where $\overleftarrow{f}$ and $\overrightarrow{f}$ 
are two gated recurrent units (GRU) proposed by~\newcite{cho+:2014_gru}.

Given the encoded ${\bf h}$, the decoder predicts the target sentence
by maximizing the conditional log-probability of the 
correct output ${\bf y^*} = (y^*_1, ... y^*_m)$, where 
$m$ is the length of the target. At each time $t$, 
the probability of each word $y_t$ from a target vocabulary $V_y$ is:
\begin{equation}
\label{eq:py}
p(y_t|{\bf h}, y^*_{t-1}..y^*_1) = g(s_t, y^*_{t-1}, H_{t}),
\end{equation}
where $g$ is 
a two layer feed-forward neural network 
over the embedding of the previous target word $y^*_{t-1}$,  
the hidden state $s_t$, and the weighted sum of ${\bf h}$ ($H_{t}$). 

To compute $s_t$ and $H_t$, we first covert $s_{t-1}$ and 
the embedding of $y^*_{t-1}$ into an intermediate state $s'_t$ with a GRU $u$ as:
\begin{equation}
s'_t = u(s_{t-1}, y^*_{t-1}).
\end{equation}
Then we have $s_t$ as:
\begin{equation}
s_t = q(s'_{t}, H_{t})
\end{equation}
where $q$ is a GRU,
and the $H_{t}$ is computed as:
\begin{equation}
H_t = 
\begin{bmatrix}
\sum_{i=1}^{l}{(\alpha_{t,i} \cdot \overleftarrow{h}_i)} \\
\sum_{i=1}^{l}{(\alpha_{t,i} \cdot \overrightarrow{h}_i)} \\
\end{bmatrix},
\end{equation}
The attention weights, $\alpha$ in $H_t$, are computed with a two layer feed-forward neural network $r$:
\begin{equation}
\alpha_{t,i} = \frac{\exp\{r(s'_{t}, h_{i})\}}{\sum_{j=1}^{l}{\exp\{r(s'_{t}, h_{j})\}}}
\end{equation}

\section{Unsupervised Approach}
\label{sec:unsup_approach}

Our unsupervised model is based on the same training idea as a denoising auto-encoder (DAE) similar to \newcite{vincent2008extracting}.
The original DAEs were feedforward nets applied to (image) data.
In our experiments, the model architecture is a seq2seq model similar to ~\newcite{bahdanau+:2014}. 
The idea of a DAE is to train a model that is able to reconstruct each training example from a partially destroyed input.
This is done by first corrupting each training sequence $x_i$ to get a partially destroyed version $\tilde{x_i}$.

In our unsupervised experiments, we
generate the training data with the following corrupting process, parameterized by the desired percentage $p$ of deletion:
for each target sequence $x_i$, a fixed percentage $p$ of words are removed at random, while the others are left untouched.
We sample a new corrupt version $\tilde{x_i}$ in each training epoch. Instead of always removing a fixed percentage of words,
we sample $p$ for each sequence separately from a Gaussian distribution with mean $p=0.6$ and variance $0.1$.
We chose $p=0.6$ based on the average length ratio between the slot values and the target sequences
in our labeled training data.

\begin{table*}[ht]
    \centering
    \begin{tabular}{c|l|l}
        & original & Loch Fyne is a family friendly restaurant providing Indian food . \\ \hline \hline
        (a) & remove random 60\%  & Fyne is restaurant food . \\ \hline
        \multirow{2}{*}{(b)} & remove only words $w_i$ & \multirow{2}{*}{Loch Fyne family friendly Indian} \\
         & with $N(w_i) > 100$ & \\ \hline
        (c) & shuffle words & family friendly Indian Loch Fyne 
    \end{tabular}
    \caption{Training data generation heuristics. (a): random 60\% of the words are removed.
    (b): 60\% of the words are removed, but only words that occur more than 100 times in the training data. Our assumption is that these are the non-content words.
    (c): On top of (b), all words are shuffled while keeping all word pairs (e.g. Loch Fyne) together that also occur in the original sentence.} 
    \label{tab:training_example_gen}
\end{table*}

This corruption approach is motivated by the fact that many NLG problems are facing a similar task to the one the DAE is solving.
Given some structured information, the task is to generate a target sequence that includes all the information.
If we map the structured information to phrases that should be in the desired output,
then the structured data problem resembles the DAE problem.
For instance, if we have the following structured example:
\emph{name: Aromi} - \emph{family friendly: yes} $\rightarrow$ \emph{Aromi has a family friendly atmosphere.}
\noindent, we convert it into the input \emph{Aromi family friendly} that we can feed to the DAE.
To preprocess the structured data, we convert the boolean feature \emph{family friendly} into a meaningful phrase ("family friendly") by using the slot name.
For all non boolean slot pairs, we just use the slot values as meaningful phrases.
Please keep in mind this transformation is only needed during inference as the training data has no slot pairs and only consists of pairs of corrupt and correct target sequences.

Nevertheless, there are two major differences between the training procedure of a DAE and an inference instance in NLG:
First, we do not  need to predict any content information in NLG as all of the content information is already provided by the structured data.
On the other hand, a DAE training instance can also remove content words from the sentence.
To align the two much closer, we restrict the words which the DAE is allowed to remove and apply the following heuristic to the corruption process of the DAE: 
Given the absolute counts $N(v_i)$ for each word $v_i$ in our vocabulary, we only allow $v_i$ to be removed when its count $N(v_i)$ is larger than a threshold.
This heuristic is motivated by the fact that the corpus frequency of content words like a restaurant name is most likely low and the corpus frequency of non-content words like "the" is most likely high.
The corpus frequencies can be either calculated on the training data itself or on a different corpora. The latter one has the advantage that domain specific content words
that are frequent in the training data will have a low frequency in an out-of-domain corpora.

The second difference is that in a DAE training instance, the words in a corrupt input occur in the same order as in the desired target.
For an NLG inference instance, the order of the structured input does not need to match the order of the words in the output. To overcome this issue, we shuffle 
the words within the corrupt sentence while not splitting bigrams that also exist in the original sentence.
An example of all three heuristics is given In Table \ref{tab:training_example_gen}.

\section{Supervised Approach}
\label{sec:sup_approach}
\begin{figure}[h]
    \centering
    \includegraphics{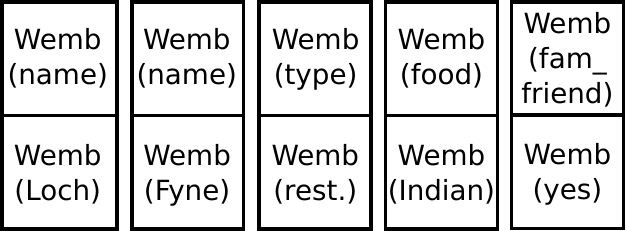}
    \caption{Example input word embeddings for our supervised baseline (Section~\ref{sec:sup_approach}) from the training Example of Table~\ref{tab:training_example}. The upper half of the word embedding is used for the slot names; the lower half for the slot values.}
    \label{fig:sup_wemb}
\end{figure}

For comparison, we train a supervised baseline based on the vanilla seq2seq model as described in Section~\ref{sec:network}.
To make better use of the structured data, we found that the input word embeddings (wemb) of the seq2seq network should be represented together by the slot name and value.
We split the word embedding vector into two parts and use the upper half for a word embedding of the slot name and the lower half for the word embedding of the slot value.
If a slot value has multiple words, we build separate word embeddings for each word, but all having the same upper part (slot name).
An example for the slot pairs of Table~\ref{tab:training_example} is given in Figure~\ref{fig:sup_wemb}.

\section{Data Sets}
\label{sec:data}

The \textbf{E2E data set}~\cite{novikova2017e2e} contains reviews in the restaurant domain. Given up to 8 different pieces of information about a restaurant, the task is to generate an English sentence that includes all of the provided structured information.
The dataset comes with 42061 training examples. We split the provided dev set (547 examples) into validation (tail 268 examples) and test (head 279 examples), each having between 3 and 42 (on average 8) reference sentences.
An example of an E2E training instance is given in Table~\ref{tab:training_example}.

The \textbf{news-commentary} data set is a parallel corpus of news
provided by the WMT conference~\cite{bojar2017findings} for training machine translation (MT) systems.
For our unsupervised experiments, we use the English \textbf{news-commentary} part of the corpora only which contains 256,715 sentences. 

All corpora are tokenized and we remove sentences that are longer than 60 tokens. 
In addition to tokenization, we also apply byte-pair encoding \cite{sennrich2016neural} when \textbf{news-commentary} is included in the setup.

\begin{table}[h]
    \centering
    \begin{tabular}{|l|l|c|}
        \hline
        dataset & task & \# examples \\ \hline \hline
        E2E & NLG & 42,061 \\ \hline
        news-commentary & MT (De-En) & 256,715 \\ \hline
    \end{tabular}
    \caption{Training data statistics. Each training instance in NLG contains structured information and one reference sequence. Each training instance in MT contains one sentence written in both German and English.}
    \label{tab:training_data}
\end{table}

\section{Experiments}
\label{sec:experiments}

\subsection{Model Parameters}
For all of our experiments we utilize the seq2seq implementation as described in Section~\ref{sec:network}. We run inference with a beam size of 5.
We use a hidden layer size of 1024 and a word embedding size of 620 and use SGD with an initial learning rate of 0.5. We halve the learning rate every other epoch starting from the 5th epoch.
We evaluate the generated text with \BLEU~\cite{papineni2002bleu},
\ROUGE~\cite{lin2004rouge}, and \NIST~\cite{doddington2002automatic} and use the evaluation tool provided by the E2E organizers to calculate the scores. 

\begin{table*}[t!]
    \centering
    \begin{tabular}{|l|l|c|c|c|}
        \hline
       setup & model & \BLEU & \ROUGE & \NIST \\ \hline \hline
        \multirow{2}{*}{supervised}     & baseline E2E challenge \cite{duvsek2016sequence} & 70.2 & 72.4 & 8.3 \\ \cline{2-5}
                                        & baseline vanilla seq2seq (Section~\ref{sec:sup_approach}) & 72.7 & 75.1 & 8.3 \\ \hline \hline
        \multirow{5}{*}{\shortstack{unsupervised}}                        & baseline copy input & 27.7 & 56.4 & 3.2 \\ \cline{2-5}
                                                             & randomly drop & 57.3 & 65.9 & 7.3 \\ \cline{2-5}
                                                             &\ \ + only words w/ count \textgreater 100 (on ind data) & 59.5 & 66.4 & 7.3 \\ \cline{2-5}
                                                             &\ \ + only words w/ count \textgreater 100 (on ood data) & 62.5 & 67.2 & 7.5 \\ \cline{2-5}
                                                             &\ \ \ \ + shuffle pos & 64.8 & 69.4 & 7.6 \\ \cline{2-5}
                                                             &\ \ \ \ \ \ + ood data & 66.0 & 71.1 & 7.7\\ \hline \hline
                                                          semi-supervised  & \ \ \ \ \ \ \ \ + labeled ind data & 67.0 & 72.1 & 7.8 \\ \hline
    \end{tabular}
    \caption{Results on the E2E dataset. The terms ood and ind are abbreviations for out-of-domain and in-domain respectively. 
    The baseline systems as well as the unsupervised systems that are not labeled with \emph{+ ood data} are trained on the E2E in-domain training data only.}
    \label{tab:e2e-results}
\end{table*}

\subsection{Automatic Scores}
Our experimental results are summarized in Table~\ref{tab:e2e-results}. We list two supervised baselines: The first one is from the organizers of the E2E challenge, the second one is from our supervised setup (Section~\ref{sec:sup_approach}). 
Our baseline yields better performance on \BLEU and \ROUGE while reaching similar performance in \NIST. 
Our third (unsupervised) \emph{baseline copy input} just runs the evaluation metrics on the input (slot values of the structured data).
This system performs much worse, but serves as a lower bound for our unsupervised experiments.

We report results on different unsupervised setups as described in Section~\ref{sec:unsup_approach}.
The system \emph{randomly drop} just randomly drops 60\% of the words, but still yields 57.3 \BLEU, 65.9 \ROUGE and 7.3 \NIST points.
You can easily detect a lot of extra information in the output that can not be explained by the structured input.
Further, the output sounds very machine generated as the output depends on the order of the structured data.
The heuristics \emph{+ only words w/ count \textgreater 100 (on ind data)} and \emph{+ only words w/ count \textgreater 100 (on ood data)} forbid removing words in the corruption phase that appear less than 100 times in the in-domain data or out-of-domain (ood) data, respectively. 
The latter setup uses the out-of-domain data for generating the word counts only and yields an improvement of 5.2 \BLEU, 1.3 \ROUGE and 0.2 \NIST points compared to just randomly dropping words.
The output still sounds very machine generated, but stops hallucinating additional information.
We further improve the performance by 3.5 \BLEU, 3.9 \ROUGE, and 0.2 \NIST points when shuffling the words in the corrupted input and using the out-of-domain data also as training examples.
We use the English side of the 256,715 sentences from the news-commentary dataset as out-of-domain data only. We did not see any further improvements by adding more out-of-domain
training data.

Finally, we build a semi supervised system that in addition to the unlabeled data includes the labeled information for some of the training examples. 
For these, we remove the slot names from the structured data and use a concatenation of all slot values as input to learn the correct output.
By jointly using both unlabeled and labeled data, we yield an additional improvement of 1.0 \BLEU points compared to our best fully unsupervised system.
In our semi supervised setup, we only use the slot values as input even for the labeled examples. This explains the drop in performance when comparing 
to the supervised setups. All supervised setups also include the slot names in their input representation.

\subsection{Human Evaluation}
\label{sec:human_eval}
In addition to automatic scores, we ran human assessment of the generated text as none of the automatic metrics correlates well with human judgment~\cite{belz2006comparing}.

To collect human rankings, we presented 3 outputs generated by 3 different systems side-by-side to crowd-workers, who
were asked to score each sentence on a 6-point scale for:

\begin{table*}[t!]
    \centering
    \begin{tabular}{|l|c|c|c|}
        \hline
        \multirow{2}{*}{system} & \multirow{2}{*}{fluency} & all & extra/ false \\
        & & information & information \\ \hline \hline
        baseline E2E challenge \cite{duvsek2016sequence} & 4.01 & 4.89 & 0.05 \\ \hline
        baseline vanilla seq2seq (Section~\ref{sec:sup_approach}) & 4.46 & 4.91 & 0.08 \\ \hline
        \shortstack{unsupervised} (random drop + words w/ count \textgreater 100 (ood data) & \multirow{2}{*}{$4.70^{\dag}$} & \multirow{2}{*}{$5.00^{\dag}$} &\multirow{2}{*}{0.05} \\
        + shuffle pos + ood data) & & & \\ \hline
    \end{tabular}
    \caption{Human evaluation results: We generated 279 output sequences for each of the 3 listed systems. Each sequence has been evaluated by 3 different raters and the score is the average of 837 ratings per system. For each task and sequence, the raters where asked to give a score between 0 and 5. A score of 5 for \emph{fluency} means that the text is fluent and grammatical correct. A score of 5 for \emph{all information} means that all information from the structured data is mentioned. A score of 0 for \emph{extra/ false information} means that no information besides the structured data is mentioned in the sequence. Scores labeled with $\dag$ are significant better than all other systems ($p < 0.0001$).}
    \label{tab:human_results}
\end{table*}

\begin{itemize}
    \item \textbf{fluency}:
How do you judge the overall naturalness of the utterance in terms of its grammatical correctness and fluency?
\end{itemize}

For the next questions, we presented in addition to the 3 different system outputs, the structured representations of each example.
We asked the crowd-worker to score the following two questions on a 5-point scale:

\begin{itemize}
    \item \textbf{all information}:
How much of the given information is mentioned in the text?
    \item \textbf{bad/ false information}:
How much false or extra information is mentioned in the text?
\end{itemize}

Each task has been given to three different raters. Consequently, each output has a separate score for each question that is the average of 3 different ratings.
The human evaluation results are summarized in Table~\ref{tab:human_results}. We included the two supervised baselines and our best unsupervised setup in the human evaluation.
The unsupervised setup outperforms the supervised setups in fluency. One explanation is that our unsupervised system includes additional unlabeled data that can not
be included in a supervised setup. Due to our unsupervised learning approach that all words in the structured data need to be included in the final output, the unsupervised
system did not miss any information. Further, all three outputs included little false or wrong information that was not included in the structured data. All in all the output
of the unsupervised system is better than the two supervised systems.

We used approximate randomization (AR) as our significance test, as recommended by \cite{riezler2005some}. Pairwise tests between
results in Table~\ref{tab:human_results} showed that our novel unsupervised approach is significantly better than both baselines regarding
fluency and mentioning all information with the likelihood of incorrectly rejecting the null hypothesis of $p < 0.0001$.

\section{Limitations}
\label{sec:limitation}
Our unsupervised approach has two limitations and is therefore not easily applicable to all NLG problems or datasets.
First, we can only run our approach for datasets where the input meaning representation either overlaps with target texts or we need to
generate rules that map the structured data to target words. Unfortunately, the needed pattern can be very complicated and the effort
of writing rules can be similar to the one of building a template based system.

Second, to be able to generate text from structured data during inference, the original structured input is converted to an unstructured one by discarding the slot names.
This can be problematic in scenarios where the slot name itself contributes to the meaning representation, but the slot name should not be in the target text.
For instances the structured data of a WEBNLG \cite{gardent2017creating} training example consists of several subject-predicate-object tuple features. Many of the features for one example have
the same subject, but different predicates and objects. But yet in the final output, we prefer to have the subject only once.

\section{Related Work}
\label{sec:related_work}

\subsection{Neural Language Generation}
Due to the recent success in Deep Learning, researchers started to use end-to-end systems to jointly model the traditional separated tasks of content selection, sentence planning and surface realization in one system.
Recently, RNNs~\cite{wen2015semantically},  attention-based methods \cite{mei2016talk} or LSTMs \cite{wen2015stochastic} were successfully applied for the task of NLG.
\newcite{DBLP:journals/corr/abs-1711-09724} introduced a modified LSTM that adds a field gate into the LSTM to incorporate the structured data.
Further, they used a dual attention mechanism that combines attention of both the slot names and the actual slot content.
\newcite{2017arXiv170900155S} extended this approach and integrated a linked matrix in their model that learns the desired order of the slots in the target text.
Further, \newcite{duvsek2016sequence} reranked the n-best output from a seq2seq model to penalize sentences that miss required information or add irrelevant ones.
Instead of RNNs, \newcite{lebret-grangier-auli:2016:EMNLP2016} introduced a neural feed-forward language model conditioned on both the full structured data and the structured information of the previous generated words.
In addition, the authors introduced a copy mechanism for boosting the words given by the structured data.

In contrast to the above mentioned related work, we train our model in a fully unsupervised fashion. Although, all our experiments have been conducted with the seq2seq model,
our unsupervised approach can be applied on top of all of the different network architectures that are introduced by the above mentioned papers.

\subsection{DAE and Unsupervised Learning}
Denoising auto-encoders and unsupervised training have been applied to various other NLP tasks.
\newcite{vincent2008extracting} introduced denoising one-layer auto-encoders that are optimized to reconstruct input data from random corruption.
The outputs of the intermediate layers of these denoisers are then used as input features for subsequent learning tasks such as supervised classification \cite{lee2009unsupervised,glorot2011domain}.
They showed that transforming data into DAE representations (as a pre-training or initialization step) gives more robust (supervised) classification performance.
\newcite{lample2018unsupervised} used a denoising auto-encoder to build an unsupervised Machine Translation model.
\newcite{hill2016learning} trained a denoising auto-encoder on a seq2seq network architecture for training sentence and paragraph representations from the output of the intermediate layers.
They showed that using noise in the encoder step is helpful to learn a better sentence representation.

In contrast to the above mentioned related work, we train a DAE directly on a task and do not take the intermediate hidden states of a DAE as sentence representation to help learning a different task.
Further, none of the related work applied DAEs on the task of generating sentences out of structured data. In addition, we modify the original DAE corruption process by introducing heuristics that
remove non-content words only to match the input representation of a supervised NLG training instance.

\section{Conclusion}
\label{sec:conclusion}
We showed how to train a denoising auto-encoder that is able to generate correct English sentences from structured data.
By applying several heuristics to the corruption phase of the auto-encoder, we reach better performance compared to two fully supervised systems.
As no labeled data is needed in our approach, we further successfully improve the quality by incorporating out-of-domain data into the training phase.
We run a human evaluation for the two supervised baselines and our best unsupervised setup. We see that the output of our unsupervised setup not only includes
100\% of the structured information, but also outperforms both supervised baselines in terms of fluency and grammatical correctness.

The unsupervised training scheme gives us the option to incorporate any unlabeled data. One possible addition to our approach would be to
incorporate text in different languages into our system, so that we can generate the output in any language from the same structured data.

Our approach is appropriate only for NLG problems where the goal is to include all the information from the structured data in the output. In future work, we will 
focus on the semi-supervised approach to make the DAE also suitable for problems where instead of all, only a subset of the structured information
should be included in the output.

\bibliography{unsup_nlg}
\bibliographystyle{acl_natbib.bst}

\end{document}